\documentclass[10pt, twocolumn]{article}
\usepackage{usenix}
\usepackage[cm]{fullpage}
\usepackage{epsfig,endnotes}
\usepackage{listings}
\usepackage[utf8]{inputenc}
\usepackage{subfigure}
\usepackage{graphicx}
\usepackage{amsmath}
\usepackage{booktabs}
\usepackage{multirow}
\usepackage{makecell}
\usepackage{siunitx}
\usepackage{xspace}
\usepackage{soul}
\usepackage[symbol]{footmisc}
\usepackage{hyperref}
\hypersetup{pdftex,colorlinks=true,allcolors=black,pdfpagemode=UseOutlines}

\DeclareFixedFont{\ttb}{T1}{txtt}{bx}{n}{8} 
\DeclareFixedFont{\ttm}{T1}{txtt}{m}{n}{8} 

\usepackage{color}
\definecolor{deepblue}{rgb}{0,0,0.5}
\definecolor{deepred}{rgb}{0.6,0,0}
\definecolor{deepgreen}{rgb}{0,0.5,0}
\definecolor{darkgreen}{rgb}{0,0.5,0}
\definecolor{purple}{rgb}{0.5,0,0.5}
\definecolor{gray}{rgb}{0.33,0.33,0.33}

\newcommand\pythonstyle{\lstset{
language=Python,
basicstyle=\ttm,
commentstyle=\color{gray},
keywords={def,return,self,with,as,for,in},
keywordstyle=\ttb\color{deepblue},
emph={TreeLSTM},                 
emphstyle=\ttb\color{deepred},    
stringstyle=\color{deepgreen},
frame=tb,                         
showstringspaces=false,
numbers=left,
xleftmargin=2em,
framexleftmargin=1.5em
}}

\lstnewenvironment{python}[1][]
{
\pythonstyle
\lstset{#1}
}
{}

\newcommand{\op}{\texttt{InvokeOp}\xspace}
\newcommand{\ops}{\texttt{InvokeOp}s\xspace}
\newcommand{\graph}{\texttt{SubGraph}\xspace}
\newcommand{\graphs}{\texttt{SubGraph}s\xspace}

\newcommand{\eat}[1]{}

\eat{
\copyrightyear{2018}
\acmYear{2018}
\setcopyright{acmlicensed}
\acmConference[EuroSys '18]{Thirteenth EuroSys Conference 2018}{April 23--26, 2018}{Porto, Portugal}
\acmBooktitle{EuroSys '18: Thirteenth EuroSys Conference 2018, April 23--26, 2018, Porto, Portugal}
\acmPrice{15.00}
\acmDOI{10.1145/3190508.3190530}
\acmISBN{978-1-4503-5584-1/18/04}
}

\begin{document}
\lstset{language=Python}   

\renewcommand*{\thefootnote}{\fnsymbol{footnote}}

\Urlmuskip=0mu plus 1mu

\title{\LARGE Improving the Expressiveness of \protect\\ Deep Learning Frameworks with Recursion\footnotemark[2]}

\author{\rm{Eunji Jeong\footnotemark[1], Joo Seong Jeong\footnotemark[1],
        Soojeong Kim, Gyeong-In Yu, Byung-Gon Chun\footnotemark[3]} \\
  Seoul National University \\
        \{ejjeong, joosjeong, soojeong\_kim, gyeongin, bgchun\}@snu.ac.kr
}

\maketitle

\footnotetext{$\dagger$ Appeared in EuroSys '18}
\footnotetext{* Both authors contributed equally to the paper}
\footnotetext{$\ddagger$ Corresponding author}

\abstract{

Recursive neural networks have widely been used by researchers to handle applications with recursively or hierarchically structured data.
However, embedded control flow deep learning frameworks such as TensorFlow, Theano, Caffe2, and MXNet fail to efficiently represent and execute such neural networks, due to lack of support for recursion.
In this paper, we add recursion to the programming model of existing frameworks by complementing their design with recursive execution of dataflow graphs as well as additional APIs for recursive definitions.
Unlike iterative implementations, which can only understand the topological index of each node in recursive data structures,
our recursive implementation is able to exploit the recursive relationships between nodes for efficient execution based on parallel computation.
We present an implementation on TensorFlow and evaluation results with various recursive neural network models,
showing that our recursive implementation not only conveys the recursive nature of recursive neural networks better than other implementations, but also uses given resources more effectively to reduce training and inference time.

}

\renewcommand*{\thefootnote}{\arabic{footnote}}
\setcounter{footnote}{0}

\section{Introduction} \label{intro}
Recursive neural networks have widely been used by researchers to handle applications with recursively or hierarchically structured data, such as natural language processing~\cite{treelstm, treernn, bowman2016fast} and scene parsing~\cite{sharma2014recursive, treernn, sharma2015deep, lin2016deep}.

In order to implement such models, \textit{embedded control flow deep learning frameworks} (in short, \textit{embedded control flow frameworks}), such as TensorFlow~\cite{tensorflow}, Theano~\cite{theano}, Caffe2~\cite{caffe2}, and MX\-Net~\cite{mxnet}, embed control flows within dataflow graphs,
i.e., the control flow is represented as a type of operation of the dataflow graph, which can trigger conditional execution or iterative computation.
However, the programming model proposed by such frameworks fails to efficiently represent and execute neural networks with recursive structures.
The designs of these frameworks do not consider recursive models and instead urge users to either write their models with iterative constructs~\cite{tf-ctrl} or completely unroll models without exploiting control flow at all~\cite{tf-padding, mxnet-padding}.
Meanwhile, \textit{non-embedded control flow deep learning frameworks} (in short, \textit{non-embedded control flow frameworks}) such as PyTorch~\cite{pytorch} or DyNet~\cite{dynet} allow users to define control flows from the client-side, creating new computation graphs for all possible control flow paths of a model. This approach trades performance for programmability, losing optimization opportunities because each graph is usually executed only once.

An important example of recursive neural networks is the Tree\-LSTM~\cite{treelstm} model, a tree-shaped network with recursively definable nodes, demanding complicated execution mechanisms.
In existing frameworks, the TreeLSTM network is handled by either statically unrolling the full network graph beforehand~\cite{pytorch, dynet}, or using a single LSTM cell to iteratively compute all intermediate nodes~\cite{tensorflow, theano}.
For the former case, it is difficult to process multiple data instances together because the tree structure differs for each instance.
For the latter case, the iterative execution is inherently sequential and thus is incapable of computing multiple nodes in parallel.

In this paper, we introduce \textit{recursive} definitions into the programming model of existing embedded control flow frameworks~\cite{tensorflow, caffe2, mxnet, theano}, adding first-class support for recursion.
By allowing users to directly express recursive definitions in application code with enhanced programmability, models with recursive data structures such as trees or graphs can be written without requiring users to use a separate complex API to express the control flow~\cite{fold}.
Also, optimization opportunities can be exploited to boost performance, such as concurrently executing child nodes in tree structures that have no dependencies between each other.

We make recursive definitions possible by introducing a special graph operation, \op, that abstracts the execution of a \graph.
Users can incorporate recursion in models by invoking a \graph within the \op that abstracts the same \graph.
The framework handles the execution of an \op as the initiation of a new \graph containing a bundle of inner operations, which are treated the same as the original running operations.

We implemented support for recursively defined dataflow graphs on TensorFlow~\cite{tensorflow}, a widely used deep learning (DL) framework.
To show the expressive power and the performance of recursive graphs, we implemented three applications using our framework:
sentiment analysis with the TreeRNN~\cite{treernn}, RNTN~\cite{rntn}, and Tree\-LSTM~\cite{treelstm} models.
For every model, we succeeded in capturing the recursive semantics of the computation graph, and achieved competitive performance compared to other state-of-the-art deep learning frameworks such as TensorFlow~\cite{tensorflow} and PyTorch~\cite{pytorch}.

The rest of the paper is organized as follows.
Section~\ref{motiv} explains the limitations of existing embedded control flow frameworks regarding recursive models, and Section~\ref{iface} provides a high-level API for efficiently representing such recursive models.
Section~\ref{design} describes the design aspects of our framework, and Section~\ref{impl} presents the implementation details.
Section~\ref{eval} presents evaluation results on various applications.
Section~\ref{rltwk} covers related work and Section~\ref{conc} concludes.

\section{Motivation} \label{motiv}

\subsection{Embedded Control Flow Frameworks and Their Limitations}

Modern deep learning frameworks use directed acyclic graphs (DAGs) to represent mathematical computations of deep learning applications and the execution order of such computations.
The vertices of graphs represent the mathematical operations, while the edges represent the dependencies between two operations.
An edge from operation \textit{a} to operation \textit{b} implies that the output of \textit{a} is fed into \textit{b} as the input value.
As the execution order between any two operations in the computation graph is statically determined, it is a non-trivial task to represent dynamic control flow within computations, such as conditionally executing only a part of the graph, or jumping to a nonadjacent operation.

Based on how to handle dynamic control flow, we can divide deep learning frameworks into two categories: \textit{embedded control flow} frameworks and \textit{non-embedded control flow} frameworks.
Embedded control flow frameworks such as TensorFlow~\cite{tensorflow} and Theano~\cite{theano} include control flow concepts inside the computation graph.
They define special kinds of control flow operations to embed the control flow within the graph.
This way, a single computation graph is able to express multiple control flow paths.
Since these frameworks can build a single graph and execute it repeatedly, aggressive performance optimization can be done while hiding the optimization overhead.

On the other hand, non-embedded control flow frameworks including PyTorch~\cite{pytorch}, DyNet~\cite{dynet}, and Chainer~\cite{chainer} do not represent the control flow inside the computation graph.
Instead, they create a new static computation graph for every activated control flow.
This approach enables fast prototyping and easy development of various deep neural networks.
However, this approach leaves little room to optimize the performance of computation graph execution, because each graph gets executed only once.

\textbf{Embedded control flow frameworks.}
In embedded control flow frameworks, graph vertices represent not only arithmetic operations (e.g., \texttt{Add} or \texttt{MatMul}) and data transformations (e.g., \texttt{Concat}), but also data-dependent control flow mechanisms.
Conditional expressions are often made available by many embedded control flow frameworks.
A predicate is expected as the first input argument, and two other operation groups as the \texttt{true} and \texttt{false} inputs.
Based on the predicate value, only one of the two operation groups are executed and passed to the output operation.
Another useful control flow construct in existing deep learning frameworks is the iterative loop construct, namely the \texttt{while\_loop} operation in TensorFlow and the \texttt{Scan} operator in Theano.
This kind of API enables adding a group of operations, referred to as a loop body, to be executed multiple times iteratively.
Conditional expressions are usually used with loop constructs to denote the termination condition of the loop body.

By planting dynamic control flow inside the computation graph and thus decoupling the client-side code execution from computation graph execution,
frameworks can exploit parallelism while executing jobs by handling mutually independent operations in a concurrent manner,
and can also exploit graph optimization techniques for faster execution that would otherwise be impossible for non-embedded control flow frameworks.
This paper will focus on embedded control flow frameworks, building up on the provided optimizations to produce maximum performance.

\textbf{Limitations of embedded control flow frameworks.}
The computation graphs of embedded control flow frameworks do not fully cover every possible control flow construct, however.
Designing recursive neural networks efficiently using embedded control flow of iterative loop constructs is difficult.
Not only is it unclear how to parallelize independent operations with iterative loops,
recursion and iteration are fundamentally different and thus converting one into another involves a nontrivial conversion process~\cite{liu1999recursion, friedman2008essentials, filinski1994recursion}.
The following subsection shows an example demonstrating the difficulties of designing recursive neural networks with just loop constructs.

\subsection{Example: TreeLSTM} \label{subsec:treelstm}
The long short-term memory~\cite{lstm} (LSTM) cell is a block of functions that is well-known for its ability to ``remember'' past computations of a neural network, and is often used for networks that process data of sequential characteristics such as text data with sentence structures.

TreeLSTM~\cite{treelstm} is a widely used recursive neural network based on LSTMs that is known to perform well for applications with tree-shaped data instances such as parse trees for natural language processing and hierarchical scene parsing~\cite{treernn}.
In an ordinary linear recursive neural network, LSTM cells are placed sequentially regardless of the input data structure.
On the other hand, in the TreeLSTM model, LSTM cells are combined to form a tree, mimicking the shape of the input data tree.
Sentiment analysis is often used as an application of the TreeLSTM.
For example, with movie review sentences and the corresponding ratings as training input data and labels, the TreeLSTM network can be trained to predict the sentiment of each movie review sentence.

There are two approaches to implement this TreeLSTM network with current deep learning frameworks, both having its own limitations.

The first approach is \textit{unrolling} the whole tree structure to the computation graph, so that LSTM cells are duplicated for each tree node.
To train multiple trees with this approach, however, a new graph must be created for all input training instances.
Not only does this result in an excessive amount of graph objects and significant construction overhead, the effect of compile-time graph optimization is near zero as all graphs are used only once.

The second approach is using \textit{iterative} control flow operations provided by frameworks.
Figure~\ref{fig:treelstm-iter} shows pseudocode of an iterative implementation of the TreeLSTM model.
In this implementation, a single LSTM cell can be used multiple times for multiple input data instances.
After the leaf nodes are processed sequentially in Line 14, the internal nodes with their dependencies resolved get processed in Line 15.
In order for this approach to work, the input tree must be preprocessed so that its nodes are assigned with topologically sorted indices, i.e., executing the tree nodes in an iterative manner does not violate the computational dependencies.
Since the recursive nature of the tree structure cannot be directly represented by iteration, it is difficult to write and understand this code.

The process of topologically sorting the tree nodes loses the parent-child node relationships of the tree, and thus the iterative implementation can only view the tree nodes as a linearly ordered list.
A recursive formulation, on the other hand, would be able to utilize the information on parent-child relationships to concurrently execute nodes, and is inherently more suitable for representing recursive neural networks, preserving their recursive nature.

\begin{figure}
\begin{python}
states = array()

def compute_leaf(idx):
  curr_state = lstm(embed(tree.leaves[idx]))
  states.insert(idx, curr_state)

def compute_internal(idx):
  left_idx, right_idx = tree.children[idx]
  left_state = states.get(left_idx)
  right_state = states.get(right_idx)
  curr_state = lstm(left_state, right_state)
  states.insert(idx, curr_state)

for_loop(range(num_leaves), compute_leaf)
for_loop(range(num_internals), compute_internal)

root_state = states[root_idx]
\end{python}
\caption{Iterative implementation of the TreeLSTM model in pseudocode.}
\label{fig:treelstm-iter}
\end{figure}

\subsection{Recursion in Embedded Control Flow Frameworks} \label{subsec:recur}

The drawbacks of the unrolling method and the iterative method suggest the need for a more effective and intuitive solution to implement TreeLSTMs, and recursive neural networks in general.
We propose that recursively defining and executing recursive neural networks is a simple yet powerful approach. 

Recursive execution of computation graphs has many similarities with recursive invocation of functions in general programming languages.
Recursive function invocation in programming languages is supported by allowing a function to call itself inside the function body.
This is usually more complicated than executing non-recursive functions, since when parsing the source code of a recursive function, the recursive function call must be processed before the parsing of the function gets finished.

Inspired by the concept of functions and function invocations, we propose to design similar ideas in embedded control flow frameworks to support recursive execution.
First, a programming interface for defining a subset of the computation graph that will be executed recursively is required.
Then, an invocation operation inside the graph subset is also needed, to trigger the recursive execution of the graph subset.
No modern embedded control flow framework supports these functionalities and, at the same time, is able to train a recursive neural network, to the best of our knowledge.

Our observations above suggest that an implementation of recursion, for embedded control flow frameworks, must satisfy two conditions.
First, recursion must be expressible as part of a valid computation graph.
Despite the fact that recursion implies the usage of a call stack of arbitrary length, the graph representation of recursion must be finite and executable by the framework.
The graph representation of recursion corresponds to the recursive \textit{function} definition; the definition simply denotes what computation is involved and how the recursion occurs, while not actually running the function.
Moreover, this representation must be usable together with other non-recursive parts of the computation graph (Section~\ref{subsec:subgraph}).

Second, an operation included in a recursive computation graph must be able to trigger the surrounding computation graph recursively.
The operation triggering the recursive graph execution corresponds to the function \textit{invocation}, which can further unfold the computation until the recursion termination condition is satisfied (Section~\ref{subsec:callop}).

\section{Programming Model} \label{iface}

In this section, we describe our modifications to the programming model of existing embedded control flow frameworks, as well as how they are translated into dataflow graph components.

\subsection{Unit of Recursion: \texttt{SubGraph}} \label{subsec:subgraph}

It is infeasible to implement the dynamism and recurrences of recursive computations using the static components of dataflow graphs provided by existing embedded control flow frameworks.
In response to this shortcoming, we first propose an abstraction, \texttt{SubGraph}, that represents basic recursive blocks and, at the same time, can be used in conjunction with existing operations to create a dataflow graph with recursive computations.

\texttt{SubGraph}s are created by grouping operations of a given computation graph that will be executed recursively.
\texttt{SubGraph}s represent fractions of a dataflow graph.
Executing a \texttt{SubGraph} refers to executing the operations that belong to that \texttt{SubGraph}.
The inputs and outputs of operations that are connected to outer operations (operations that reside outside of the current \texttt{SubGraph}) are assigned as inputs and outputs of the \texttt{SubGraph} itself.
During execution, the inputs of a \texttt{SubGraph} are passed to the corresponding inner operations, while operation outputs that must be shipped out to outer operations are passed as \texttt{SubGraph} outputs.
A \texttt{SubGraph} can be regarded as a function in general programming languages.

Additionally, we allow \texttt{SubGraph}s to invoke other \texttt{Sub\-Graph}s.
A \texttt{SubGraph} invocation within an outer \graph is connected to the other inner operations to form an inner dataflow graph, just as the outer \texttt{SubGraph} is connected to outer operations.
A \texttt{SubGraph} invocation in a \texttt{SubGraph} simply implies that there is yet another group of operations to be executed at that particular graph position.
Coming back to the function analogy, placing a \texttt{SubGraph} invocation within a \texttt{SubGraph} is identical to calling a function within another function.

More importantly, a \texttt{SubGraph} may recursively invoke itself.
This aspect makes possible the definition of a recursive computation;
we define a recursive block as a \texttt{SubGraph} and insert a invocation to itself in the same \texttt{SubGraph}.

Figure~\ref{fig:treelstm-recur} shows the recursive implementation of the Tree\-LSTM model, with details omitted for brevity.
After defining a \texttt{SubGraph} for the TreeLSTM model in Line 2, we reuse the definition in Lines 10-11 to complete the recursive tree structure of the model.
Notice how a conditional expression is used (\texttt{if} in Line 14) to separate the base case from the recursive case.
Comparing with Figure~\ref{fig:treelstm-iter}, this recursive version follows the definition of the TreeLSTM model more clearly; the recursive nature of the tree structure is explicitly represented in this implementation.

\begin{figure}[h] 
\begin{python}
# TreeLSTM: index(int32) -> hidden_state(Tensor)
with SubGraph() as TreeLSTM:
  idx = TreeLSTM.input(int32)

  def compute_leaf_node():
    return LSTM(embed(tree.leaves[idx]))

  def compute_internal_node():
    left_idx, right_idx = tree.children[idx]
    left_state = TreeLSTM(left_idx)
    right_state = TreeLSTM(right_idx)
    return LSTM(left_state, right_state)

  TreeLSTM.output(if(is_leaf_node(idx),
                     compute_leaf_node,
                     compute_internal_node))

root_state = TreeLSTM(root_idx)
\end{python}
\caption{Recursive implementation of the TreeLSTM model with \texttt{SubGraph} definitions.
After declaring the start of a \texttt{SubGraph} in Line 2, we indicate the inputs of the \texttt{SubGraph} in Line 3.
The body of the \texttt{SubGraph} is defined in Lines 5-16, while recursive calls are made on Lines 10-11.
Note that \texttt{SubGraph} outputs must be given as in Lines 14-16.
The completed \texttt{SubGraph} definition can now be used in Line 18.
}
\label{fig:treelstm-recur}
\end{figure}

\subsection{Recursion in Dataflow Graphs: \op} \label{subsec:callop}

While \texttt{SubGraph}s provide a convenient way to define recursive computations, the framework is still left with the task of actually executing the operations gathered as \texttt{SubGraph}s.
However, as \texttt{SubGraph} operations are expected to be executed in a recursive fashion, an additional mechanism for ``re-executing'' the operations of \texttt{SubGraph}s repeatedly (until some termination condition is met) is required.
To this end, we introduce a new operation named \op.

An \op is an operation that takes a set of tensors as input, runs an associated \texttt{SubGraph} (i.e., executes the inner operations of the \texttt{SubGraph}) with the provided inputs, and returns the resulting tensors as output.
\ops are execution objects instantiated from \texttt{SubGraph} invocations; as \texttt{SubGraph}s are semantically close to function definitions, \ops can be considered as function calls to the functions specified by \texttt{SubGraph}s.
As such, it is possible for a single \texttt{SubGraph} to be associated with more than one \op.

Despite the special property of having an associated \texttt{SubGraph}, an \op is fundamentally the same as other operations such as \texttt{Add} or \texttt{MatMul}, and is generally treated as an ordinary operation.
The difference with other operations comes from the operation kernel implementation; instead of performing a mathematical calculation, an \op abstracts the execution of an entire \texttt{SubGraph}.
This difference also affects a process called automatic differentiation, a core functionality provided by modern deep learning frameworks for training models.
Instead of calculating mathematical derivates of some numerical function like other operations, the framework must take into account the associated \graph of an \op.
We will discuss this further in Section~\ref{subsubsec:backprop-rec}.

\subsection{TreeLSTM with \texttt{SubGraph}s \& \ops}
\begin{figure}[tbp]
  \centering
  \subfigure[]{
    \includegraphics[width=0.99\columnwidth]{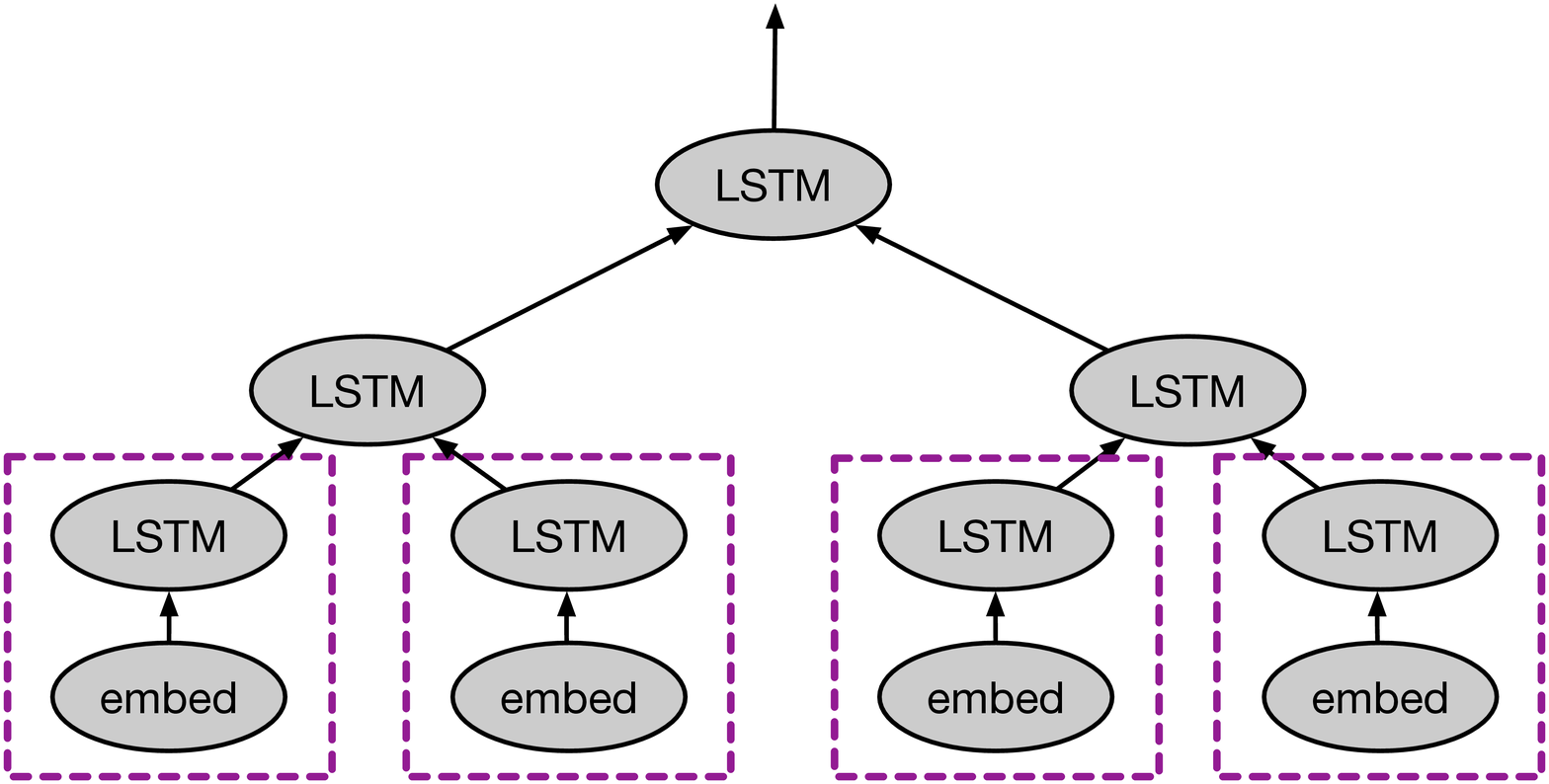}
  }
  \vfill
  \subfigure[]{
    \includegraphics[width=0.99\columnwidth]{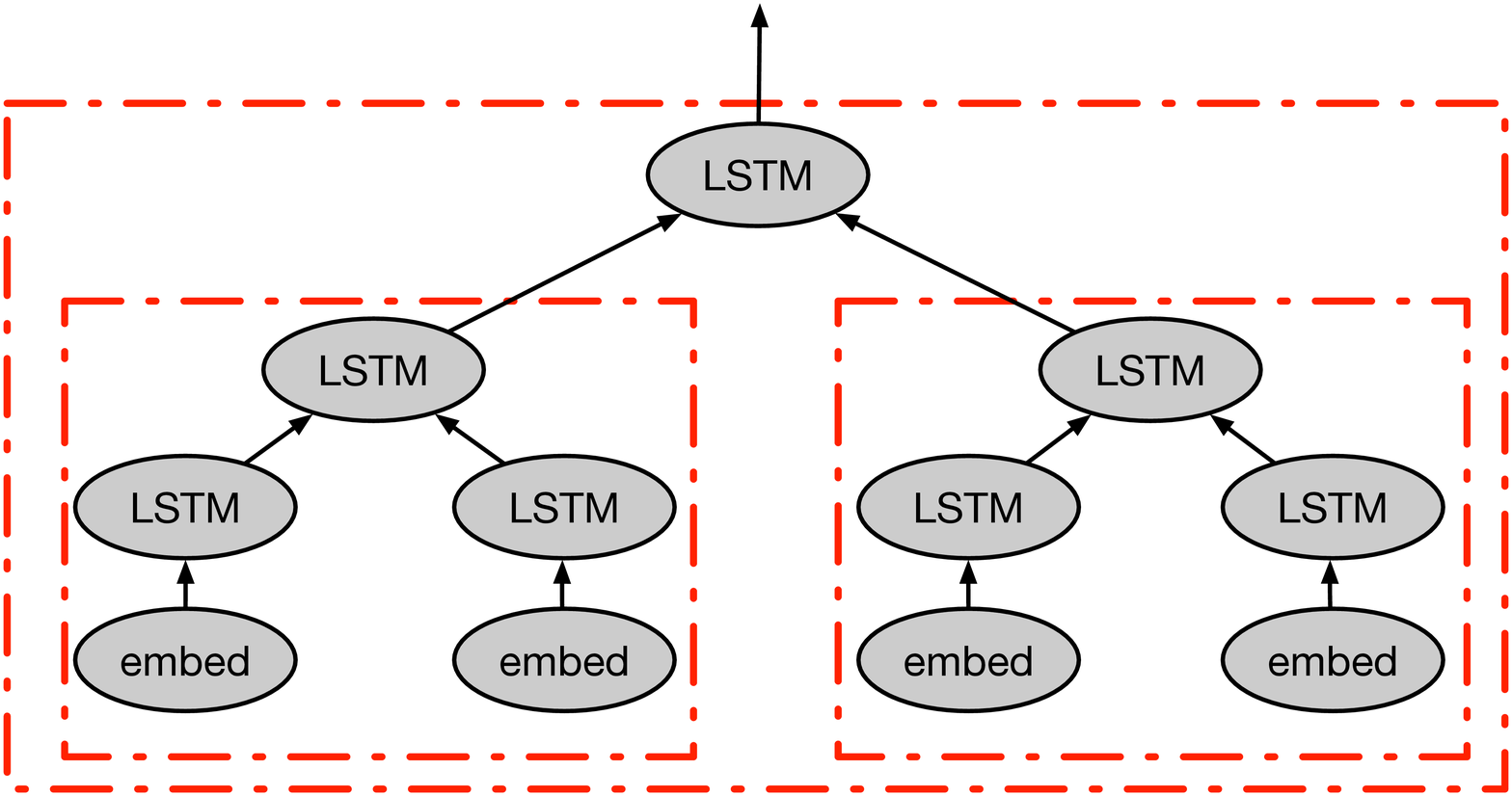}
  }
  \vfill
  \subfigure[]{
    \includegraphics[width=0.99\columnwidth]{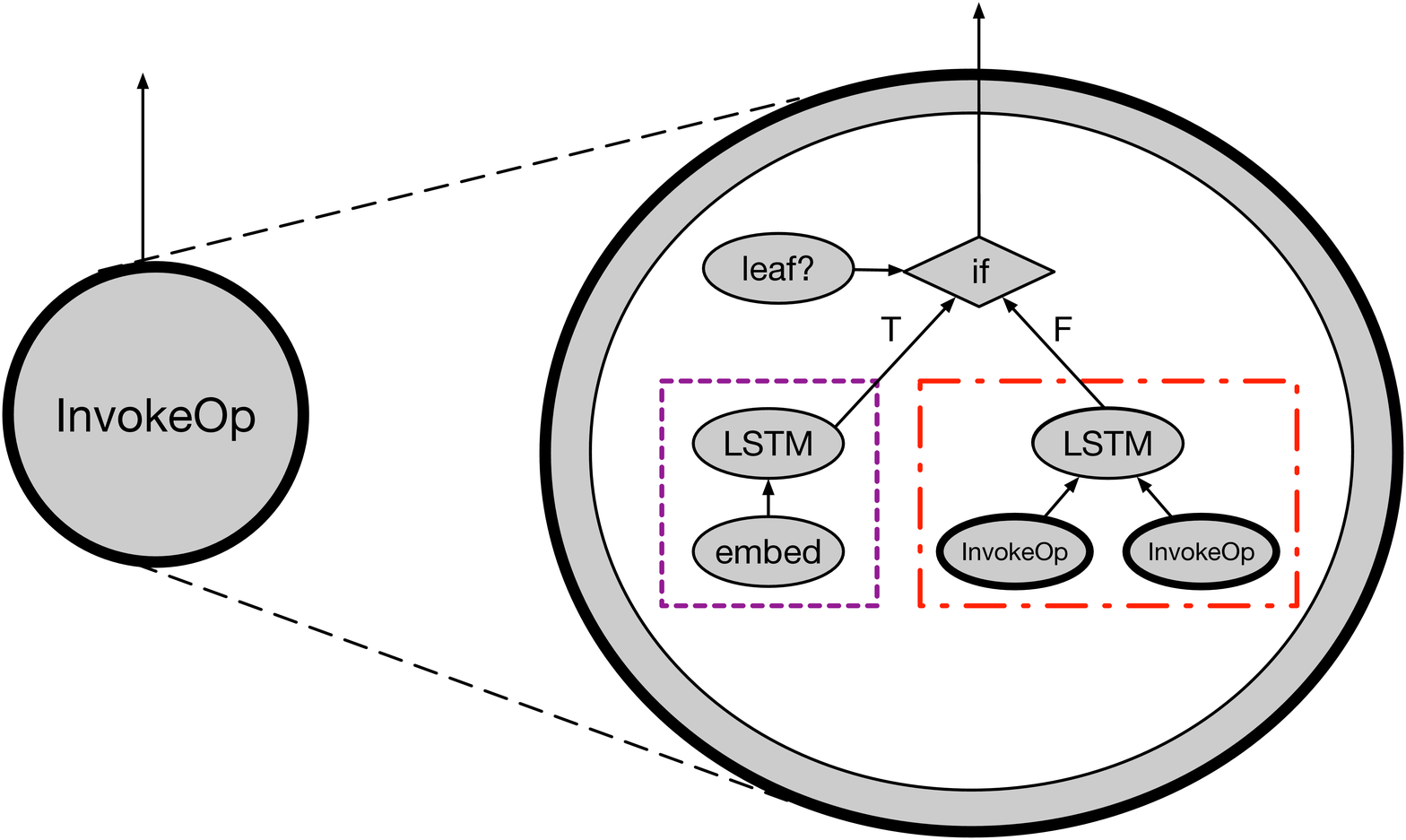}
  }
  \caption{
	  An illustration of how an unrolled computation graph of the TreeLSTM model (a, b) can be transformed into a recursive graph with \ops (c).
    The base case, depicted in the boxes of (a), and the recursive case, indicated by the boxes in (b), can be combined to succinctly describe the model as a recursive \texttt{SubGraph} as shown in (c).
    \ops have been added at the appropriate places to mark the points where a recursive function call to the \texttt{SubGraph} must occur.}
  \label{fig:treelstm-callop}
\end{figure}

Figure~\ref{fig:treelstm-callop} portrays an example on how \ops are used to represent the TreeLSTM (Section~\ref{subsec:treelstm}) model with recursion.
A completely unrolled depiction of the model for a full binary tree is shown in Figures~\ref{fig:treelstm-callop}(a) and \ref{fig:treelstm-callop}(b).
It is not hard to observe that the model can be expressed using recursion: the \texttt{embed} operation and the \texttt{LSTM} cell at the leaves form the base case (Figure~\ref{fig:treelstm-callop}(a)), while the two-input \texttt{LSTM} cell at the intermediate notes corresponds to the recursive case (Figure~\ref{fig:treelstm-callop}(b)).

Merging the base case and the recursive case into a \texttt{SubGraph} with a conditional branch (\texttt{if}), we now have a concise representation of the TreeLSTM model, as shown in Figure~\ref{fig:treelstm-callop}(c).
Note that the condensed \texttt{SubGraph} is able to represent TreeLSTMs of arbitrary height or shape, and not just a single particular structure.
\ops are inserted at all inner recursive call points.

\section{System Design} \label{design}

In this section, we discuss various system design aspects for supporting the recursive programming model of the previous section.

Our design complements existing embedded control flow frameworks with additional APIs for declaring recursive graphs and core changes for executing such recursive graphs.
Models declared using the \graph API from Section~\ref{iface} are transformed into a dataflow graph containing \ops.
In turn, the framework core engine runs the resulting graph with the same mechanism used to run non-recursive graphs, accessing additional graph and value cache structures when dealing with \ops.
The design does not involve any implementation details of a particular framework, and can be applied to any DL framework that incorporates control flows in computation graphs.

\subsection{Graph Execution}

\subsubsection{Background: Execution Model of Existing Frameworks}
The execution model of embedded control flow frameworks can be characterized by three components:
the \textit{client} who builds and submits dataflow graphs to the framework,
the \textit{master} which parses the given graphs and schedules the execution of operations,
and one or more \textit{workers} that carry out the actual computation of the operations.
The master coordinates the execution of operations on the workers, running operations in an order that respects the inter-operation dependencies.

\begin{figure}[tb]
  \includegraphics[width=0.95\linewidth]{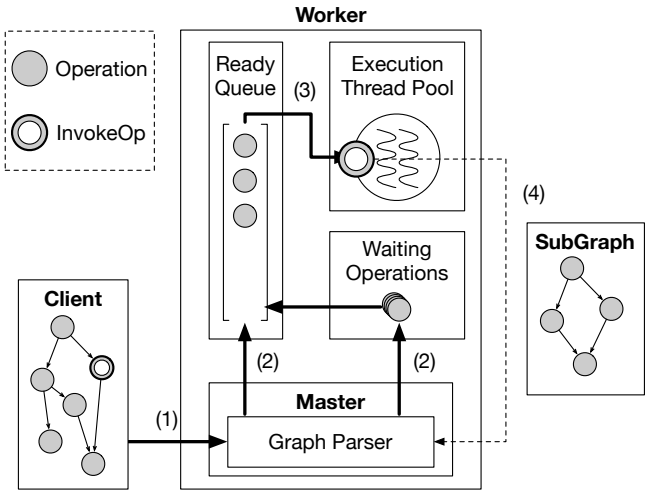}
  \centering
  \caption{The execution model of embedded control flow frameworks with \ops.
(1) After the client initiates the job with a dataflow graph, (2) the master decomposes the graph into operations and places them into either the ready queue or the waiting line of the worker, depending on the number of unresolved inputs.
(3) Operations are dequeued from the queue by idle execution threads, while new operations are enqueued when input dependencies are resolved.
(4) When an \op gets executed, its associated \graph is passed to and processed by the master, similar to step (1).
Only one worker is shown for the sake of clarity.}
  \label{fig:execmodel}
\end{figure}

Steps (1)-(3) of Figure~\ref{fig:execmodel} displays an illustration of the execution model, with only one worker shown for simplicity.
When the master first analyzes the input dataflow graph, operations that require no inputs are enqueued directly into the \textit{ready queue} of the worker, whereas operations that need at least one input are put on standby.
Next, execution threads of the worker's \textit{execution thread pool} grab operations from the operation queue and perform the computation for those operations in parallel.
When an execution thread finishes running an operation, the master checks the waiting operations that have a dependency on the completed operation, and enqueues operations whose dependencies have all been resolved to the ready queue.
This process is repeated until all operations have been processed.

\subsubsection{Recursive Execution}
The execution mechanism for executing a recursively defined dataflow graph is no different from the mechanism for executing non-recursive graphs.
This is possible because the execution of an \op mimics the initiation of a new dataflow graph, with the exception of reusing the same master scheduler as well as the same worker ready queues, as illustrated in step (4) of Figure~\ref{fig:execmodel}.
When an \op becomes executable and is dequeued by an execution thread, the graph associated with the \op is processed by the master, similar to how a graph submitted by the client is parsed by the master.
Operations that are immediately ready to be run are enqueued into the ready queue, behind the existing operations.
Likewise, operations that have at least one unresolved input dependency are added to the waiting list, together with other previous standby operations.

This design allows recursive dataflow graphs to be processed on existing embedded control flow frameworks without drastic changes.
Recursive graphs can enjoy graph optimizations supplied by such frameworks and achieve good performance while providing intuitive, recursive APIs at the same time.
In fact, from the framework's point of view, a recursive graph is the more general representation, while non-recursive graphs are simply special cases which have no recursive \graphs and \ops.

It is also worth noting that performing priority scheduling of operations instead of simple FIFO scheduling may possibly yield significant effects on the execution time of recursive computation graphs, depending on the inter-operation dependencies of the given recursive model.
For example, if the model contains a \texttt{SubGraph} whose inner operation must be processed in order for many outer operations to be enqueued into the ready queue, then a scheduling decision of processing inner operations before others would lead to a shorter execution time overall.
Although this is an interesting problem, it is usually not an issue for servers with many parallel computation threads to spare and thus we leave this as future work.

\textbf{Graph execution stack.}
When a function is invoked in programming languages, the language runtime maintains a call stack to record the call history and relevant information.
This enables the program to correctly return from the callee function to the corresponding caller function, and also provides helpful information to programmers such as backtrace information when an exception occurs while executing the function.
A similar process of keeping track of the \graph invocation history is required for the recursive graph execution engine.
However, the caller-callee relationship of \ops cannot be managed with a linear stack, because an \op can branch out into multiple child \ops in parallel.
Rather, the relationship is maintained as a tree, where each \op holds a pointer to its parent \op (i.e., return location).

\subsection{Graph Backpropagation} \label{subsec:backprop}

\subsubsection{Background: Automatic Differentiation} \label{subsubsec:backprop}
Neural networks are normally trained via the backpropagation algorithm~\cite{rumelhart1988learning}.
Backpropagation calculates the errors of all operation output values, by first comparing the final outputs of a neural network with expected ground-truth values (labels) to compute output errors, and then propagating the output errors all the way back up to the input according to the chain rule.
The calculated errors -- often referred to as \textit{gradients} -- are used to update model parameters so that operation outputs are pushed towards the expected values.

\begin{figure}[tb]
  \centering
  \subfigure[]{
    \includegraphics[width=0.8\columnwidth]{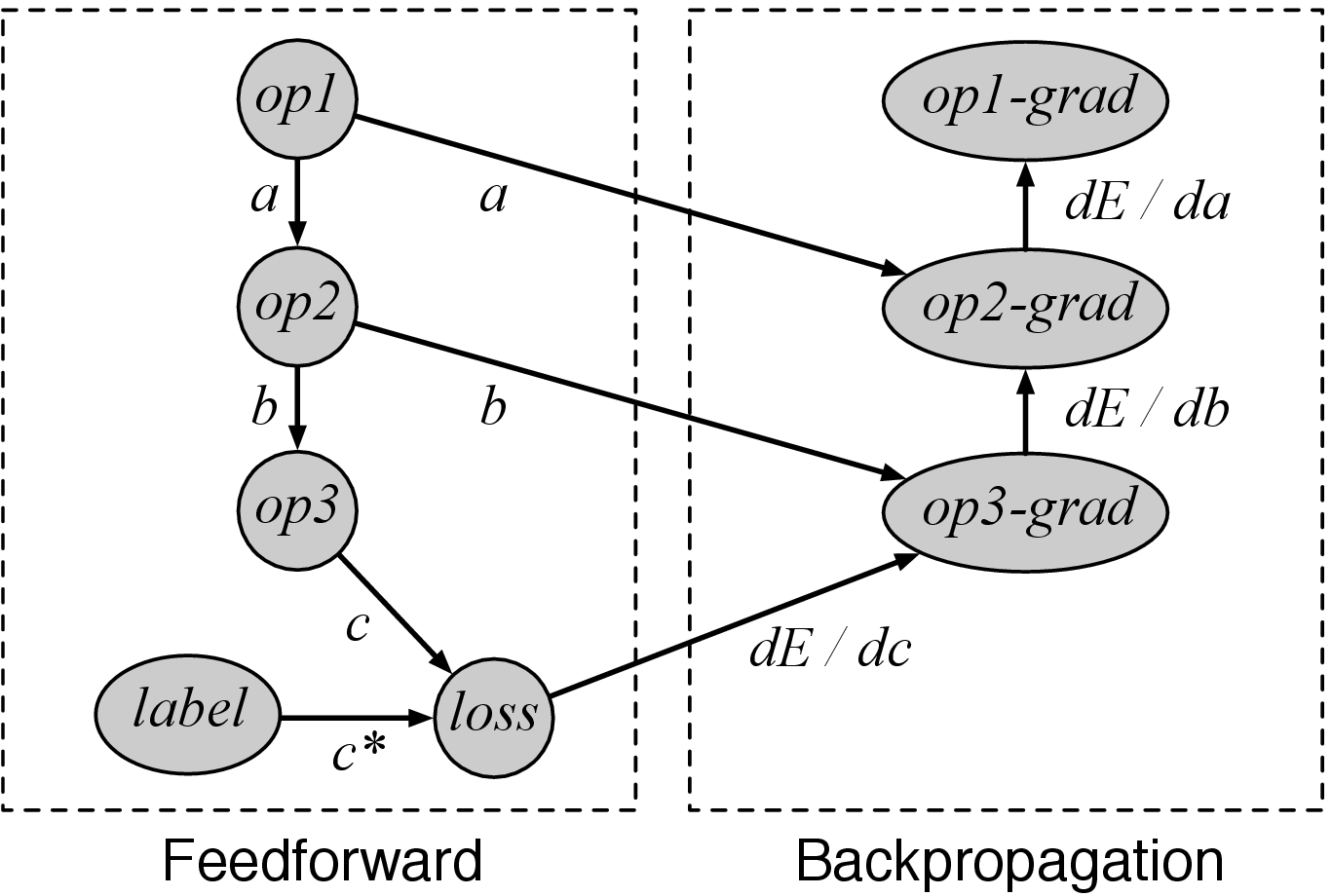}
  }
  \vfill
  \subfigure[]{
    \includegraphics[width=0.8\columnwidth]{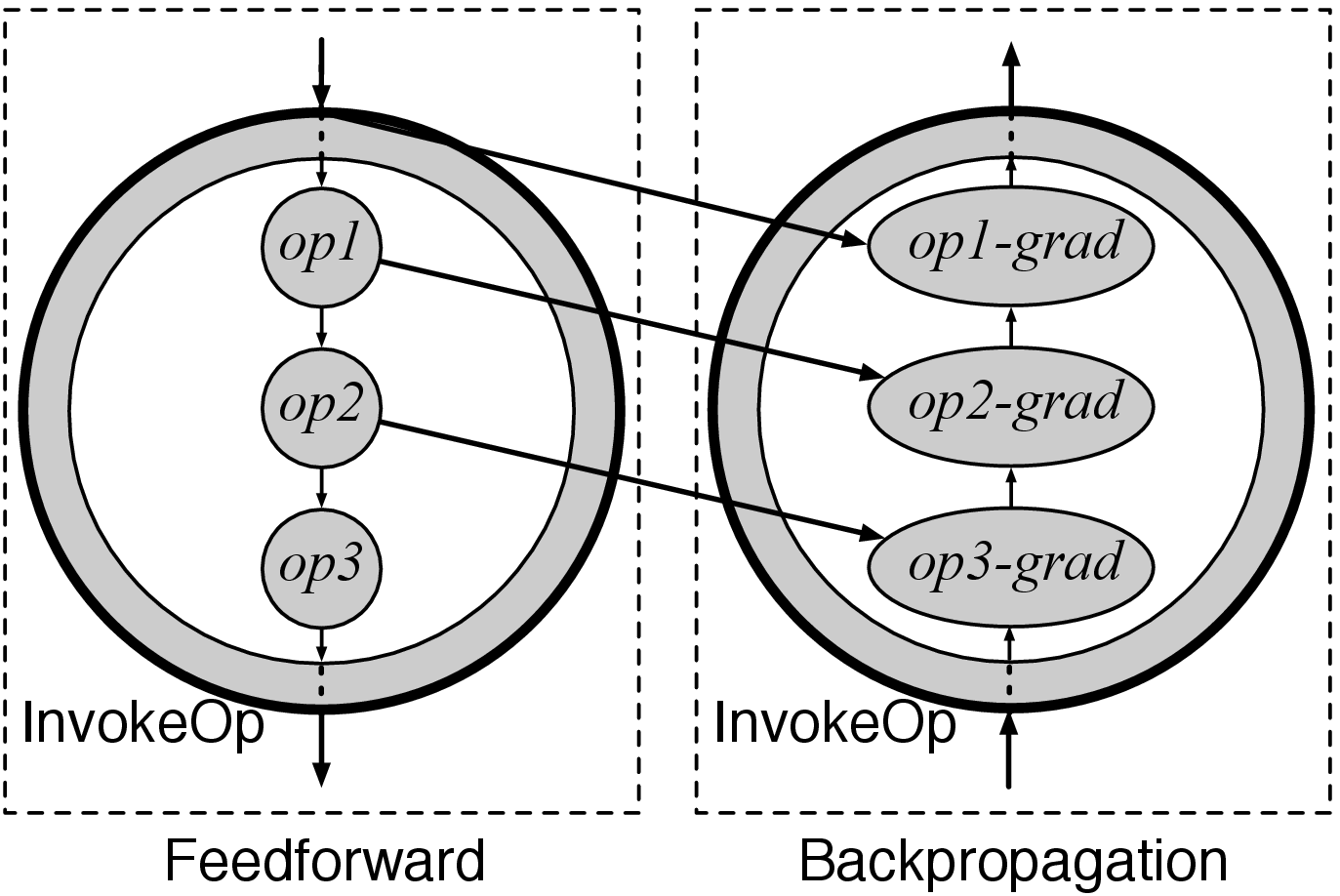}
  }      
  \caption{Backpropagation of dataflow graphs with and without \ops.
(a) A simple linear feedforward network is shown on the left, while the backpropagation side of the same network is shown on the right.
All gradient operations receive previous gradient values from its gradient predecessor as well as the original feedforward value from the feedforward network.
(b) An \op and its gradient operation for backpropagation are shown on the left and right, respectively.
Notice how (a) and (b) are structurally very similar, except for the enclosing \ops.}
  \label{fig:backprop}
\end{figure}

Backpropagation of a simple linear network is shown in Figure~\ref{fig:backprop}(a).
Starting from operation \textit{op1}, all forward operations \textit{op1}, \textit{op2}, and \textit{op3} are computed in succession to produce values \textit{a}, \textit{b}, and \textit{c}, respectively.
The final output \textit{c} is checked with the expected value \textit{c*} to produce the loss value \textit{E}, as well as the gradient of \textit{E} with respect to \textit{c}, denoted as \textit{dE/dc}.
Next, the other gradients are generated one by one, this time through the backpropagation line of operations, \textit{op3-grad}, \textit{op2-grad}, and \textit{op1-grad}.

Note that in order to calculate a certain gradient, both the previous gradient and the corresponding forward value are required.
For instance, the gradient \textit{dE/db} is computed with the previous gradient \textit{dE/dc} and the forward value \textit{b} (\textit{op3-grad}).
Likewise, \textit{dE/db} and \textit{a} are used to compute \textit{dE/da} (\textit{op2-grad}).
This results in a final dataflow graph where a forward operation shares its inputs with its backpropagation equivalent (e.g. \textit{op2} and \textit{op2-grad} both take \textit{a} as input).
As a precaution to prevent values from being invalidated (released from memory) before being consumed by all dependent operations, DL frameworks always keep all operation outputs as valid data until that particular iteration terminates.\footnote{Technically, we could recompute the forward operation values during backpropagation instead of retaining them to save memory. However, this incurs a significant increase in training time and is generally not preferred.}

\textbf{Automatic differentiation.}
Deep learning frameworks relieve users from the burden of manually inserting backpropagation operations, with the help of a process called automatic differentiation.
In the case of embedded control flow frameworks, after a user submits a feedforward neural network to the framework, the framework automatically adds all operations required for computing gradients to the given dataflow graph.
Maintaining a catalogue of predefined gradient operations, the framework backtracks along the feedforward path and adds the corresponding gradient for each and every feedforward operation.
The resulting computation graph can then be processed by the framework for execution.
As setting up the backpropagation path is usually much more tedious than defining the forward path, the automatic differentiation process is very helpful for users and currently supported by all deep learning frameworks.

\subsubsection{Recursive Backpropagation} \label{subsubsec:backprop-rec}

Backpropagation of a recursive dataflow graph is similar to backpropagation of a non-recursive dataflow graph.
The only nontrivial issue is how to define and calculate gradients for \ops.
As the feedforward output of an \op is the execution output of its associated \texttt{SubGraph}, naturally the gradient of an \op is also generated from the gradients of the associated \texttt{SubGraph}.

During automatic differentiation, we inspect the \texttt{Sub\-Graph}s associated with \ops and perform automatic differentiation on them as well.
For each \texttt{SubGraph}, we collect the gradient operations that were inserted by automatic differentiation.
At this point, it is possible to simply add the inserted gradient operations to the backpropagation path of the computation graph.
However, in case the \texttt{SubGraph} was used for recursion, the gradients for the inner recursive computations would not be generated and thus backpropagation would be returning incorrect results.

Instead, we wrap each set of gradient operations from \texttt{SubGraph}s with yet another \texttt{SubGraph}.
If a feedforward \texttt{SubGraph} contains a recursive invocation to itself, then its corresponding backpropagation \texttt{SubGraph} will also hold a recursive invocation, at the same position.
Later, \ops are inserted at \graph invocation points for both the feedforward \texttt{SubGraph} and the backpropagation \texttt{SubGraph} to complete the computation graph.

Figure~\ref{fig:backprop}(b) illustrates how a gradient operation of an \op is formed.
The associated \texttt{SubGraph} is shown in the inner side of the feedforward \op, while the corresponding gradient operations of the \texttt{SubGraph} are shown inside the backpropagation \op.
Carrying over operation outputs from the feedforward phase to the backpropagation phase is done by connecting the outputs and inputs of the relevant operations, same as in Section~\ref{subsubsec:backprop}.

\section{Implementation} \label{impl}
We implemented our framework atop TensorFlow~\cite{tensorflow} 1.3.0.
Framework changes, including the kernel implementation of \op as well as internal data structures, were done in the C++ core, while client-side APIs for recursive definitions are exposed in Python.
Here, we describe several implementation issues of our framework.

\textbf{Forward declaration.}
In embedded control flow frameworks, all operations must have well-defined inputs and outputs before they are used (comparable to function signatures in programming languages).
\ops are not exceptions; the framework does not allow the creation of a recursive \op unless the operation definition for the recursive call is specified beforehand.
This rule can be bypassed by using forward declarations for \ops that are recursively defined; when a \graph is defined, we first predeclare an empty \op that has the same signature as the \graph, and later ``register'' the \graph definition to the empty \op.
Note that this procedure is automatically done by the framework, and is not exposed to users.
Gradients for backpropagation are defined in a similar manner, with the operation declaration coming before the actual definition.

\textbf{Backpropagation cache implementation.}
As described in Section~\ref{subsec:backprop}, operation output values from the feedforward phase must be retained until backpropagation and be fed into the corresponding gradient operations.
For non-recursive computation graphs, embedded control flow frameworks would accomplish this simply by holding a feedforward value entry for each required operation and later passing the values to the appropriate gradient operations.
Unfortunately, for recursive graphs, an operation within a \graph may be called more than one time across multiple recursions.
Multiple output values generated during multiple executions must all be passed to the corresponding gradient operation, without losing their topological position information.

\begin{figure}[tb]
  \centering
    \includegraphics[clip,width=1.0\columnwidth]{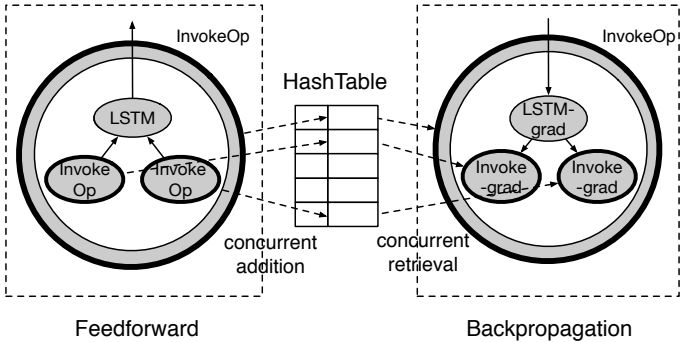}
    \caption{Concurrent hash table being used between multiple forward and backward executions of the same operation (\op).}
  \label{fig:backprop-impl}
\end{figure}

We resolve this issue by maintaining a concurrent hash table for storing and fetching operation output values of \graphs.
Figure~\ref{fig:backprop-impl} describes the whole procedure of passing multiple feedforward output values from \ops.
A hash table is externally generated and managed per \graph, and a unique hash entry key is used to distinguish table entries.
During the feedforward phase, we store all output values of \op instances that come from the \texttt{SubGraph} in the table.
An \op's key is defined by combining the \op's topological position within the \graph with the key of the parent \op, guaranteeing uniqueness.
By using a concurrent hash table, multiple instances of the same operation in the graph can concurrently access and update the hash table.

Next, during backpropagation, we perform a hash table lookup for each gradient operation of the \op instances and feed the stored value as input.
This enables feedforward output values to be retained and correctly retrieved for backpropagation.
While the concurrent insert operations may incur minor overhead, the lookup operations are thread-safe and are negligible.
Is it notable that using a simple queue or stack to store activation values is inadequate, as the order of enqueue and dequeue operations or push and pop operations is not deterministic and thus output values may be directed to the wrong gradient operation. 

\textbf{Outer reference.}
It is common for a recursive \graph to not refer to the external input values explicitly, but rather implicitly.
For instance, a static value that is required for all levels of recursion of a \graph should not be included as an input of the \graph, as the value does not change anyway.
Nonetheless, the TensorFlow framework regards a \graph and the outer graph as two separate graphs and is unable to understand the identities of such implicit external values unless they are specified as actual operation inputs.
Therefore, when a \graph is created, we analyze the operations to check whether there are any external inputs that have not already been specified as the \graph's inputs and add them to the input list.

\textbf{Implementation on other frameworks.}
Recursively defined \graphs and \ops can be implemented on not only TensorFlow but any other embedded control flow DL frameworks as well, with the computation graph and the operations as its elements.
A \graph can be provided as an abstraction that is similar to the framework's original graph structure but contains only a subset of all operations to mark a recursion block.
An \op can be implemented as a new kind of user-defined operation type, which recursively executes a \graph.

For instance, Caffe2~\cite{caffe2} uses a \texttt{NetDef} protocol buffer as its computation graph, and allows feeding \texttt{NetDef}s as inputs to operators.
By extending \texttt{NetDef}s to recursively represent subgroups of operators, we can create a Caffe2 version of \op that receives such subgroups as inputs and executes them.
Theano~\cite{theano} provides the \texttt{Scan} operator which abstracts the execution of a loop in a single operator.
Although the \texttt{Scan} operator is usually used to express iterative control flow, the concept of maintaining a separate graph isolated from the main computation graph fits well with \graphs and is a good starting point for implementing recursive computations.

\section{Evaluation} \label{eval}
We evaluate our framework while focusing on the performance benefits of the recursive nature of the framework, mostly made possible by exploitation of parallelism in recursive neural networks.

\begin{figure*}[h]
  \includegraphics[width=\textwidth]{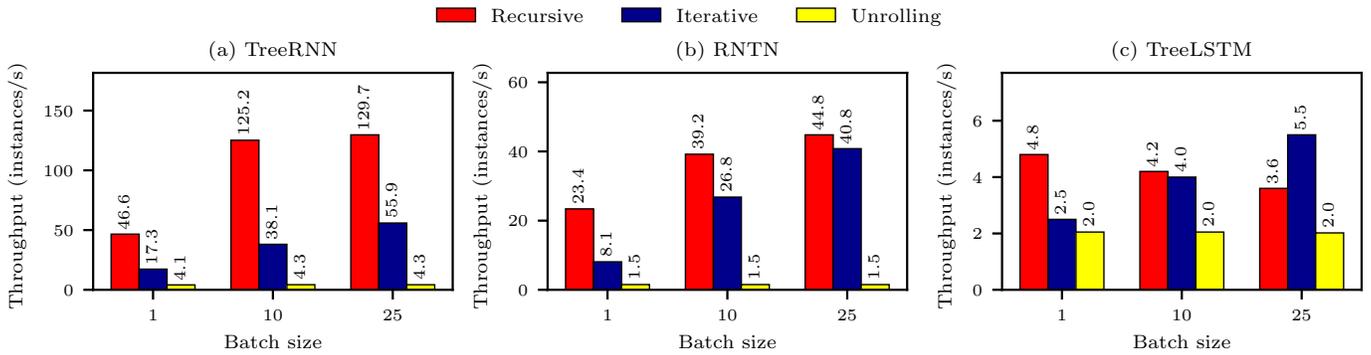}
  \centering
  \caption{
Training throughput for the TreeRNN, RNTN, and TreeLSTM models with the Large Movie Review dataset.
Numbers are shown for our recursive implementation, TensorFlow’s iterative implementation, and PyTorch's static unrolling implementation.
Our recursive implementation outperforms the other frameworks for all models and all batch sizes except when training TreeLSTM with a batch size of 25, at which point the amount of system resources is insufficient to completely parallelize the computation.
We did not observe any significant performance gain for the static unrolling approach when the batch size was increased.
}
  \label{fig:throughput-train}
\end{figure*}

\begin{figure*}[h]
  \includegraphics[width=\textwidth]{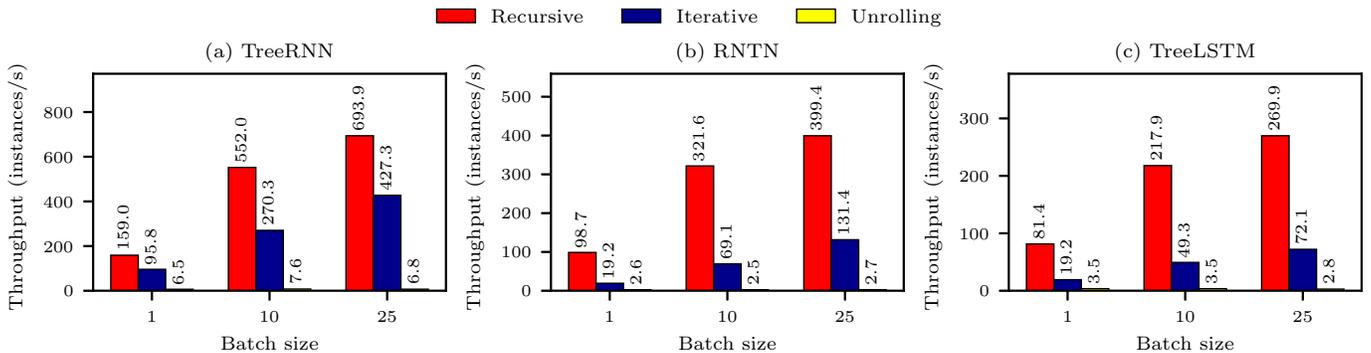}
  \centering
  \caption{
Inference throughput for the TreeRNN, RNTN, and TreeLSTM models with the Large Movie Review dataset.
Measurements are presented for our recursive implementation, TensorFlow's iterative implementation, and PyTorch's static unrolling implementation.
Our recursive implementation outperforms the other frameworks for all models and all batch sizes.
}
  \label{fig:throughput-eval}
\end{figure*}

\subsection{Experimental Setup}

\textbf{Applications.}
We implemented a variety of neural network models from the recursive neural network family, namely the aforementioned TreeLSTM~\cite{treelstm} model as well as the TreeRNN~\cite{treernn} and the RNTN~\cite{rntn} model.
All models were trained and tested to perform sentiment analysis on the Large Movie Review~\cite{wvSent} dataset, where data instances are sentences in the form of binary parse trees.
For this dataset, we used a pre-trained network (for each model) to label all nodes.
For all models, we used the same hyperparameters as the ones suggested in the original papers.
We also considered smaller batch sizes to investigate the performance trends of our framework without mixing in additional performance gains obtainable from batching instances.

\textbf{Frameworks.}
Along with our implementation of \textit{recursive} data\-flow graphs (built on TensorFlow 1.3.0), we also implemented neural networks on other frameworks, including TensorFlow~\cite{tensorflow} (TF 1.3.0) without recursive graphs, which allows an \textit{iterative} way of programming, and PyTorch~\cite{pytorch} (PTH 0.3.1), which only supports the static \textit{unrolling} technique.
Since native TensorFlow does not support recursive definitions, we used TensorFlow's control flow operators to train the neural networks in an iterative fashion, as shown in Section~\ref{motiv}.
For PyTorch, we dynamically create a new graph structure for each sentence.
Although implementing the static unrolling technique on TensorFlow is possible, the graph generation overhead tends to be very large;
instead, we use PyTorch for the unrolling technique, which incurs negligible graph construction overhead.

\textbf{Hardware specification.}
All numbers reported in this paper were retrieved from experiment results on a single machine of two 18-core Intel Xeon E5-2695 @ 2.10 GHz processors with 256GB RAM, unless otherwise specified.
We also used an NVIDIA Titan X GPU for certain models.
Unlike other common neural networks such as convolutional models, the unstructured input data of recursive neural networks makes it difficult to exploit the full computational power of GPUs.
Thus, we use GPUs only if they introduce performance gain compared to CPU-only execution.
Our implementation and TensorFlow showed greater performance on CPUs, while PyTorch performed better on a GPU.

\subsection{Throughput and Convergence Time}

We start our analysis by measuring the training and inference throughputs with the recursive, iterative, and static unrolling implementations.

\begin{figure*}[]
  \includegraphics[width=1.0\textwidth]{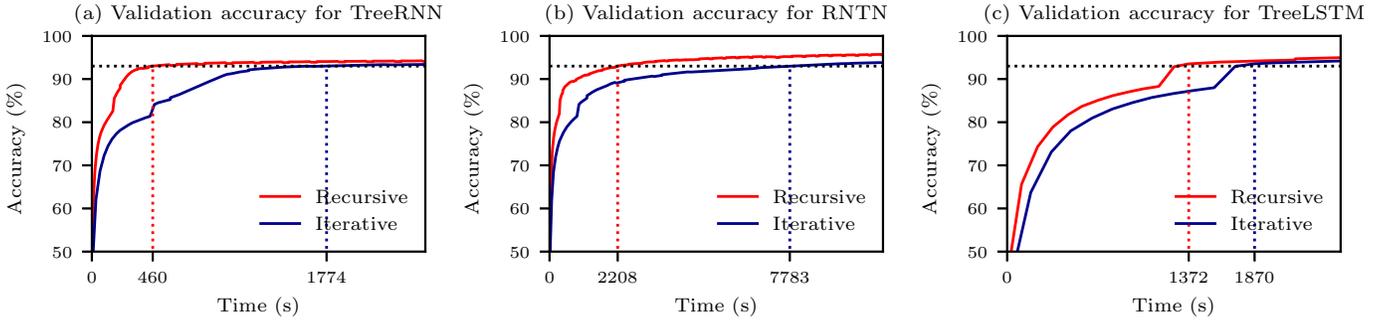}
  \centering
  \caption{
Validation accuracy for the binary sentiment classification task with (a) TreeRNN, (b) RNTN, and (c) TreeLSTM models. Results are shown for training each model with the recursive and iterative implementations, using the Large Movie Review dataset.
The time to reach 93\% accuracy for each setup is also plotted, showing that the recursive implementation converges faster for all models.
}
  \label{fig:convergence}
\end{figure*}

\textbf{Training throughput.}
Figure~\ref{fig:throughput-train} shows the throughput of training the TreeRNN, RNTN, and TreeLSTM models using the Large Movie Review dataset with recursive, iterative, and static unrolling implementations.
The models were trained with batch sizes of 1, 10, and 25.\footnote{The original TreeRNN, RNTN, and TreeLSTM papers state that using a batch size of 25 yielded the best model.}

Thanks to the parallelism exploited by recursive dataflow graphs, our implementation outperforms other implementations for the TreeRNN and RNTN models at all batch sizes, by up to 3.3x improved throughput over the iterative implementation, and 30.2x improved throughput over the static unrolling approach.
Note that the performance gap between the recursive and iterative approach for the TreeRNN model is bigger than that of the RNTN model.
This is due to the fact that the TreeRNN model involves much less computation in its recursive function body compared to the RNTN model, therefore having bigger room for performance optimization via computation parallelization.
We will further discuss the effectiveness of parallelization in Section~\ref{subsec-parall}.

For the TreeLSTM model, our implementation performs better than other frameworks at batch sizes 1 and 10.
On the other hand, at a batch size of 25, our implementation is slower than the iterative implementation.
Generally, recursion has additional overheads compared to iteration, including passing around arguments and return values, caller and callee context setup, etc.
We also have additional overheads related to backpropagation, as discussed in previous sections.
Consequently, our recursive implementation exhibits excessively high resource utilization for computing large batches, making the throughput lower than the iterative computation.

\textbf{Inference throughput.}
\textit{Inference} refers to the process of computing the operation output values of the feedforward phase, stopping before backpropagation.
Aside from training throughput, inference throughput is also a useful metric for computing the performance of a neural network,
indicating how quickly a deployed model can process unseen data, e.g., in serving systems.

Figure~\ref{fig:throughput-eval} shows the inference throughput, with identical environments with the previous experiments on training throughput.
Our framework demonstrates throughput up to 5.4x higher than the iterative implementation, and 147.9x higher than the static unrolling approach.
Unlike training throughput, our recursive implementation greatly dominates other implementations, since our framework can fully utilize the given resources and the additional overheads introduced by backpropagation are not present.

\begin{figure}[]
  \includegraphics[width=0.8\columnwidth]{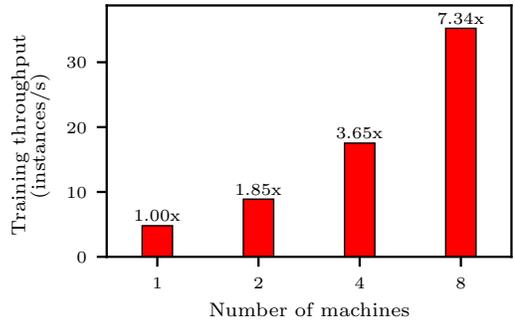}
  \centering
  \caption{
	  Training throughput for the TreeLSTM model on our recursive implementation, using varying numbers of machines for data parallelism.
	  The performance increases almost linearly as more machines are used for training.
}
  \label{fig:scalability}
\end{figure}

\textbf{Convergence.}
We also measured how the accuracy of the model increases as the training progresses, in Figure~\ref{fig:convergence}.
Since our implementation calculates numerically identical results as the iterative implementation, the accuracy improvement per epoch is the same.
However, thanks to our higher throughput, training with our framework results in faster convergence than the iterative implementation.

\textbf{Training throughput with multiple machines.}
One way to overcome the resource limitations is scaling out to multiple machines.
Figure~\ref{fig:scalability} shows how the training throughput for the Tree\-LSTM model on our recursive implementation changes, as the number of machines used in training gradually grows from 1 to 8.
Utilizing the well-known data parallelism technique~\cite{ps}, the training throughput improves almost linearly up to 8 machines.

\subsection{Analysis of Recursive Graphs: \\Parallelization} \label{subsec-parall}

The performance difference between the iterative and recursive implementation of the same recursive neural network mainly comes from the parallelization of recursive operations.
In this subsection, we analyze how the performance varies depending on various aspects related to parallelization.

\begin{figure}[]
  \includegraphics[width=1.0\columnwidth]{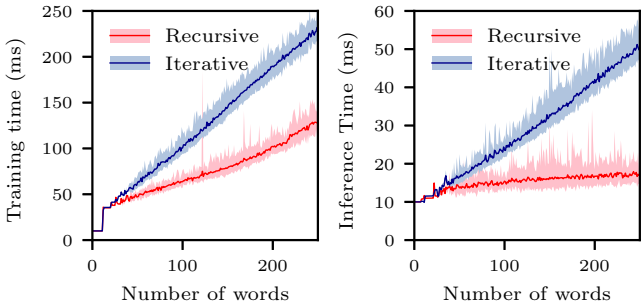}
  \centering
  \caption{
Time taken for processing each data instance, in the TreeLSTM model using the Large Movie Review dataset.
The bold lines represent the average time for each specific sentence length in the whole dataset, and the enclosing colored areas represent the range of time taken to process the specific length of sentences.
No batching is used for this experiment.
As the number of words inside a data instance increases, our recursive implementation outperforms the iterative implementation thanks to the parallelized execution of tree cells.
For inference, the computation load is low enough for the framework to utilize system resources without hitting the resource limit, and the processing time of the recursive implementation is $O(logN)$, where $N$ is the number of words.
}
  \label{fig:length}
\end{figure}

\textbf{Sentence length.}
A close inspection of the training time per data instance sorted by sentence length gives us interesting results.
As shown in Figure~\ref{fig:length}, the time required for processing a single sentence generally increases as sentences become longer, regardless of whether the implementation is based on native TensorFlow or our recursive implementation.
This is an expected phenomenon, because longer sentences form larger tree structures consisting of more cells which require more computation.

However, there is a clear difference in the increasing slope; the training time grows at a steeper slope for TensorFlow than that of our implementation.
This is because the recursive implementation allows tree cells to be processed concurrently, whereas the iterative TensorFlow implementation is only capable of processing one tree cell at a time.
Theoretically, our implementation is able to process a tree structure consisting of $N$ cells in $O(logN)$ time (native TensorFlow requires $O(N)$ time),
though the parallelization effect is diminished by the framework overhead and therefore the performance is more close to a linear trend rather than a logarithmic trend.
On inference workloads with much less resource needs, the trend is clearly closer to a logarithmic scale.

\begin{table}[tb]
\centering
  \begin{tabular}{lrrr}
    \toprule
    \multirow{2}{*}{Batch size} &
      \multicolumn{3}{c}{Throughput (instances/s)} \\
      & {Balanced} & {Moderate} & {Linear} \\
      \midrule
    1 & 46.7 & 27.3 & 7.6 \\
    10 & 125.2 & 78.2 & 22.7 \\
    25 & 129.7 & 83.1 & 45.4 \\
    \bottomrule
  \end{tabular}
  \caption{Throughput for the TreeRNN model implemented with recursive dataflow graphs, using datasets of varying tree balancedness.
    The balanced dataset exhibits highest throughput thanks to the high degree of parallelization, but at the same time does not improve as well as the linear dataset when the batch size increases from 1 to 25, because there is only a small room of performance improvement left, w.r.t parallelization.}
  \label{tbl:balancedness-treernn}
\end{table}

\textbf{Balancedness of trees.}
To analyze the influence of tree balancedness on training throughput on our recursive implementation,
we prepared several modified versions of the Large Movie Review dataset, that contain the same sentences as the original dataset but have different parse tree shapes.
Specifically, we prepared 1) a \textit{balanced} dataset consisting of only complete binary trees, 2) a \textit{moderate} dataset that contains moderately balanced binary trees, and 3) a \textit{linear} dataset comprising only extremely unbalanced binary trees.

Table~\ref{tbl:balancedness-treernn} shows the throughput of training the TreeRNN model using these three datasets.
For all batch sizes, the training throughput on the balanced dataset is the highest, while the throughput on the linear dataset is the lowest.
This trend occurs because the maximum possible execution concurrency of a tree is affected by the balancedness of the tree.
A full binary tree of $N$ cells can be processed concurrently with at most $\frac{N+1}{2}$ threads, because all $\frac{N+1}{2}$ leaf nodes are mutually independent.
On the other hand, an extremely unbalanced binary tree can be processed with only one or two threads at a time due to the linearity of the tree.
As a result, our implementation can train input data of balanced trees with greater throughput than input data of unbalanced trees.

\textbf{Resource Utilization.}
Another interesting fact in Table~\ref{tbl:balancedness-treernn} is that the training throughput on the linear dataset scales better than the throughput on the balanced dataset, as the batch size increases.
For the balanced dataset, the recursive implementation efficiently utilizes many threads to process the data even at a small batch size of 1, and thus increasing the batch size leads to a relatively small speed boost.
On the contrary, for the linear dataset, the recursive implementation fails to efficiently make use of CPU resources and thus the performance gain provided by increasing the batch size is relatively high.

\subsection{Comparison with Folding}

The performance improvement of our recursive framework discussed in previous subsections comes from executing multiple tree nodes in parallel.
On the other hand, another approach for efficiently executing recursive neural networks exists: identifying concurrently executable nodes and batching them into a single node to be run on GPUs.
We refer to this technique as \textit{folding}, following the name of a framework, TensorFlow Fold~\cite{fold}, that implements this technique.

The folding technique hardly suffers from resource limitations, as GPUs are very efficient in batching computations.
However, batching multiple nodes leads to overheads that are not present in other approaches.
Due to the various tree structures in the input data, the batching decision must be done in a depth-wise manner, thus the ungrouping and regrouping of tree nodes across multiple depths lead to numerous memory reallocations and copies.
Moreover, folding is applicable only if the tree structure of the input data is known before executing the computation;
for dynamically structured models the folding technique cannot be implemented.
Here, we discuss and compare our recursive framework with the folding technique.
Experiment results for folding were obtained using the TensorFlow Fold framework.

\begin{table}[tb]
\centering
  \begin{tabular}{lrrrrrr}
    \toprule
    \multirow{3}{*}{\makecell{Batch\\size}} &
                       \multicolumn{6}{c}{Throughput (instances/s)} \\
                     & \multicolumn{3}{c}{Inference} & \multicolumn{3}{c}{Training} \\
                       \cmidrule(lr){2-4}\cmidrule(r){5-7}
		                 & Iter  & Recur & Fold  & Iter & Recur& Fold\\
		\midrule
		1                & 19.2  & 81.4  & 16.5  & 2.5  & 4.8  & 9.0  \\
		10               & 49.3  & 217.9 & 52.2  & 4.0  & 4.2  & 37.5 \\
		25               & 72.1  & 269.9 & 61.6  & 5.5  & 3.6  & 54.7 \\
    \bottomrule
  \end{tabular}
  \caption{
		  Throughput for processing the TreeLSTM model on our recursive framework, Fold's folding technique, and TensorFlow's iterative approach, with the Large Movie Review dataset.
The recursive approach performs the best on inference with efficient parallel execution of tree nodes,
while the folding technique shows better performance on training thanks to its GPU exploitation.
}
  \label{fig:fold}
\end{table}

\subsubsection{Statically Structured Models}
Table~\ref{fig:fold} compares the throughput of performing inference and training on the TreeLSTM model using our implementation, the iterative approach, and the folding technique.
The amount of resources is sufficient for executing forward computations, and therefore our framework outperforms the folding technique for the inference task with up to 4.93x faster throughput.
Unlike folding, the recursive approach does not have any overheads regarding batch regrouping, since the calculated values can be directly passed between caller and callee \graphs.

However, when the resource usage is high, not every scheduled tree node in the worker ready queue can be executed concurrently, even if the dependencies have been fully resolved.
While the scalability of the recursive approach is limited by this drawback for the training task, the folding technique can exploit the GPU and scales better.
As a result, the folding technique performs better than the recursive approach for the training task.
We can improve the performance of the recursive approach by conditionally deciding whether to batch the operations or not similar to the folding technique, and we leave this as future work.

\subsubsection{Dynamically Structured Models}

While the models presented in the previous sections demand support for dynamic control flow, there is yet another collection of models that boast an even greater degree of dynamism, in which the model structure is gradually formed depending on intermediate values calculated from prior steps.
Top-down TreeLSTM~\cite{td-treelstm} (TD-TreeLSTM) is a dynamic model proposed for sentence completion and dependency parsing.
When a trained model receives root node information as an input, the model can generate child nodes based on the information and completes the rest of the tree sentence.
The decision of generating a child node or stopping tree expansion is conditionally made based on the computed value of the current node at runtime, so the structure of the complete tree is not known before actually executing the graph.
DRNN~\cite{drnn} is a neural network model that can generate tree-structured data from sequences, and therefore the tree structure in unknown before graph execution, similar to TD-TreeLSTM.
The Hierarchical Mixtures of Experts~\cite{hmoe, out} model has a similar structure, where the whole tree structure is decided at runtime.
The network structure of HMRNN~\cite{chung2017hierarchical} is also dynamically determined by the intermediate computation values.

\begin{table}[tb]
\centering
  \begin{tabular}{lccc}
    \toprule

    \multirow{2}{*}{Batch size} &
      \multicolumn{3}{c}{Throughput (instances/s)} \\
      & Iterative & Recursive & Folding \\

      \midrule
      \makecell[l]{1\\64}          & \makecell{0.30 \\ 0.34}   & \makecell{5.59 \\ 9.30}      &   Not supported    \\
    \bottomrule
  \end{tabular}
  \caption{
\setcounter{footnote}{2}
Throughput for evaluating the TD-TreeLSTM model on our recursive framework and TensorFlow's iterative implementation, on batch sizes of 1 and 64.\protect\footnotemark~
Being able to execute tree nodes in parallel lets our framework perform better than the iterative approach.
Fold's folding technique is inapplicable to the TD-TreeLSTM model.
}
  \label{fig:td-treelstm}
\end{table}

\footnotetext{We follow the suggestions of the original TD-TreeLSTM paper to use a batch size of 64.}

Our framework performs well for such dynamic models.
Table~\ref{fig:td-treelstm} shows the throughput of the sentence completing task with the TD-TreeLSTM model.
Our implementation performs better than the iterative approach by up to 18.6x, since multiple tree nodes are executed in parallel.
For this kind of model, techniques that rely on heavy preprocessing of input data to batch operations (folding) are ineffective because the model structures are unknown until the main computation.
We note that it is impossible to express such models using the API provided by the Fold framework.

\section{Related Work} \label{rltwk}

\textbf{Embedded control flow frameworks.}
DL frameworks with a computation graph comprised of control flow operators along with the mathematical operators to represent a DL job are called \textit{embedded control flow} frameworks~\cite{tensorflow, theano, caffe2, mxnet}.
This class of frameworks does not use the programming language's control flow support (e.g., Python's \texttt{if} clause) for representing dynamic control flow.
Instead, they provide certain primitives for embedding dynamic control flow in the dataflow graph; the framework cores evaluate a boolean expression and decide what to apply for the next operation at graph execution time.

Although our implementation is based on the embedded control flow framework TensorFlow~\cite{tensorflow},
the key difference is the ability to express recursive functions.
In our implementation, a user can define an arbitrary function and use it as an operation to compose a graph.
The arbitrary function can call another function including itself without restriction, allowing recursive definitions of functions.
TensorFlow and Theano~\cite{theano} also let users write user-defined functions, but do not support recursion; the user must not create a cycle of dependencies between functions.

\textbf{Non-embedded control flow frameworks.}
Unlike embedded control flow frameworks, PyTorch~\cite{pytorch}, DyNet~\cite{dynet}, and Chainer~\cite{chainer} do not embed control flow operators inside their computation graphs.
Rather, the computation occurs along with the dynamic control flow of the host language, removing the need to embed the control flow operators inside the computation graph.
In other words, these \textit{non-embedded control flow} frameworks behave just like numerical computation libraries such as NumPy~\cite{numpy} and MKL~\cite{mkl}, so one can directly exploit the underlying language's abilities for handling conditional branches, loops, and recursive functions.
Thanks to this behavior, a user can easily build a prototype of a new neural network architecture or optimization algorithm.

However, since neural networks are usually trained for numerous steps until they converge, non-embedded control flow frameworks suffer from repetitive construction of graphs composed of hundreds or thousands of nodes, resulting in substantial object creation and removal overhead.
More importantly, embedded control flow frameworks employ graph compilation techniques like operation fusion or in-place operation conversion to optimize execution, but non-embedded control flow frameworks cannot since they do not reuse the graphs.

Recursive dataflow graphs are designed to provide a similar level of programmability to non-embedded control flow frameworks, without losing optimization opportunities by using an embedded control flow framework (TensorFlow) to declare computations with recursion.

\textbf{Other frameworks with recursion support.}
TensorFlow Fold \cite{fold}, a library for handling models with dynamic computation, allows recursion for writing computation graphs.
Fold provides a number of new APIs for creating and managing \textit{blocks} (sets of low-level operations).
A block behaves as a scheduling unit to enable dynamic batching of different computation graphs.
Using these blocks, Fold constructs an execution loop that resembles recursion and starts running the loop from base cases, wiring intermediate results to the appropriate positions for subsequent recursive cases.
From the perspective of programmability, Fold provides a whole new set of functional programming style APIs to preprocess input data and build the computation graph.
It is required to mix the control flow API of Fold and the computational API of TensorFlow to represent a complete DL job, which is not a trivial task.
Also, since the structure and execution order of the computation graph becomes completely different after graph preprocessing, it becomes impossible to pinpoint the location of errors on failures, resulting in poor debuggability.

On the other hand, our framework adds a simple abstraction, \graph, to the programming model to support recursion.
\texttt{Sub\-Graph}s can be used with existing operations analogously and does not import any additional execution details other than those already provided by the underlying embedded control flow framework.
Moreover, the final computation graph of \ops retains the original position information of \graphs, allowing the same debugging experience as the underlying framework.

CIEL~\cite{ciel} is a dynamic task (operator) creation framework that allows users to declare data processing jobs recursively.
The operators of CIEL are relatively more coarse-grained compared to DL frameworks, which means the number of recursion calls is not large.
The different granularity comes from the characteristics of the target domain; CIEL targets batch processing applications, whereas recursively defined graphs were designed for deep learning.
More fundamentally, CIEL cannot be integrated with modern DL frameworks because CIEL does not consider DL-specific mechanisms such as backpropagation or typed operator definitions, which are highly important for DL applications.

\section{Conclusion} \label{conc}

In this paper, we have introduced recursive declarations and recursive execution mechanisms for running recursive neural networks on top of existing embedded control flow frameworks.
With recursively defined computation graphs, recursive neural networks can be implemented in a fashion that better portrays the recursion aspect, and be executed efficiently by letting the framework exploit parallel execution of computations, both of which were very difficult to achieve on existing frameworks due to the lack of support for recursion.
To achieve this goal, we designed and implemented a programming model and a runtime execution model, including automatic differentiation support for deep learning jobs.
We have demonstrated the expressive power and performance of recursive graphs by implementing various recursive neural network models using our programming model and comparing them with iterative and unrolling implementations, showing that recursive graphs outperform other approaches significantly.

\section*{Acknowledgement}
This research was supported by the MSIT (Ministry of Science and ICT), Korea, under the SW Starlab support program (IITP-2018-R0126-18-1093) supervised by the IITP (Institute for Information \& communications Technology Promotion), and by the ICT R\&D program of MSIT/IITP (No.2017-0-01772, Development of QA systems for Video Story Understanding to pass the Video Turing Test).

\bibliographystyle{abbrv}
\bibliography{eurosys18-final135}

\end{document}